\documentclass[letterpaper, 10 pt, conference]{ieeeconf}
\IEEEoverridecommandlockouts
\pdfoutput=1 % Use PDFLaTeX in arXiv. Must be placed in the first 5 lines.

\usepackage{graphicx}
\usepackage{epsfig}
\usepackage{mathptmx}
\usepackage{times}
\usepackage{amsmath}
\usepackage{amssymb}
\usepackage{diagbox}
\usepackage{url}
\usepackage[usenames]{xcolor}
\usepackage{multirow}

\DeclareRobustCommand{\bigO}{%
  \text{\usefont{OMS}{cmsy}{m}{n}O}%
}

\newcommand{\cframe}[1]{{\smash{\protect\underrightarrow{\mathcal{F}}_{#1}}}}

\title{\LARGE \bf Fast Frontier-based Information-driven \\
    Autonomous Exploration with an MAV}
\author{Anna Dai$^{1,2,*}$,
	Sotiris Papatheodorou$^{2,*}$,
	Nils Funk$^{2}$,
	Dimos Tzoumanikas$^{2}$,
	and
	Stefan Leutenegger$^{2}$%
\thanks{This work was funded in part by the President's PhD Scholarship
and in part by the EPSRC under Grant Aerial ABM EP/N018494/1.}
\thanks{$^{*}$ These authors contributed equally to this work.}
\thanks{$^{1}$ The author is with ETH Zurich.}%
\thanks{$^{2}$ These authors are with the Smart Robotics Lab, Department of
	Computing, Imperial College London. E-mail addresses: {\tt\small
	\{anna.dai19, s.papatheodorou18, nils.funk13, dimosthenis.tzoumanikas14,
	s.leutenegger\}@ic.ac.uk}}
\thanks{Video link: \protect\url{https://youtu.be/tH2VkVony38}}
}

\begin{document}
\maketitle
\thispagestyle{empty}
\pagestyle{empty}

\begin{abstract}
	Exploration and collision-free navigation through an unknown environment is
	a fundamental task for autonomous robots. In this paper, a novel
	exploration strategy for Micro Aerial Vehicles (MAVs) is presented. The
	goal of the exploration strategy is the reduction of map entropy regarding
	occupancy probabilities, which is reflected in a utility function to be
	maximised. We achieve fast and efficient exploration performance with tight
	integration between our octree-based occupancy mapping approach, frontier
	extraction, and motion planning--as a hybrid between frontier-based and
	sampling-based exploration methods. The computationally expensive frontier
	clustering employed in classic frontier-based exploration is avoided by
	exploiting the implicit grouping of frontier voxels in the underlying
	octree map representation. Candidate next-views are sampled from the map
	frontiers and are evaluated using a utility function combining map entropy
	and travel time, where the former is computed efficiently using sparse
	raycasting. These optimisations along with the targeted exploration of
	frontier-based methods result in a fast and computationally efficient
	exploration planner. The proposed method is evaluated using both simulated
	and real-world experiments, demonstrating clear advantages over
	state-of-the-art approaches.
\end{abstract}

\begin{keywords}
Aerial Systems: Perception and Autonomy, Visual-Based Navigation
\end{keywords}

\section{Introduction}
The use of MAVs in applications such as mining, outdoor large scale inspection,
precision agriculture and search and rescue have gained much popularity
recently. Fast and thorough autonomous exploration is a key factor for safe and
effective operation. Due to their agility and speed, MAVs are well suited for
mapping 3D environments.  As the exploration algorithm should be running
on-board an MAV, it should be fast and light-weight. The most common
exploration planning strategies are frontier-based and sampling-based
exploration which are compared in \cite{Julia2012comparison}. Frontier-based
methods generate their planning tasks from frontiers which are defined as
boundaries between free and known space, whereas sampling-based methods most
commonly grow Rapidly-exploring Random Trees (RRTs)
\cite{LaValle_PlanningAlgorithms2006} to compute the exploration path.

In this work, we present an exploration strategy which can achieve real-time
performance on-board an MAV. The main contributions of the strategy are as
follows:
\begin{itemize}
	\item Frontier-based exploration is combined with sampling-based
		exploration by sampling candidate next-views from the map frontiers.
		This approach achieves focused exploration like frontier-based methods,
		while requiring fewer candidate next-views than typical sampling-based
		methods due to the targeted sampling.
	\item By leveraging the implicit voxel grouping in the octree map
		representation, the computationally expensive step of frontier voxel
		clustering in classical frontier-based methods can be avoided by
		considering frontier voxels as clusters if they reside in the same
		octant of the octree.
	\item Performing a computationally expensive expected sensor measurement
		map update like conventional information-theoretic methods is avoided
		by observing that only relative ranking between candidate next-views is
		needed. Thus a much faster map entropy estimation using sparse
		raycasting is performed.
\end{itemize}

\begin{figure}[t]
    \centering
    \includegraphics[width=0.239\textwidth]{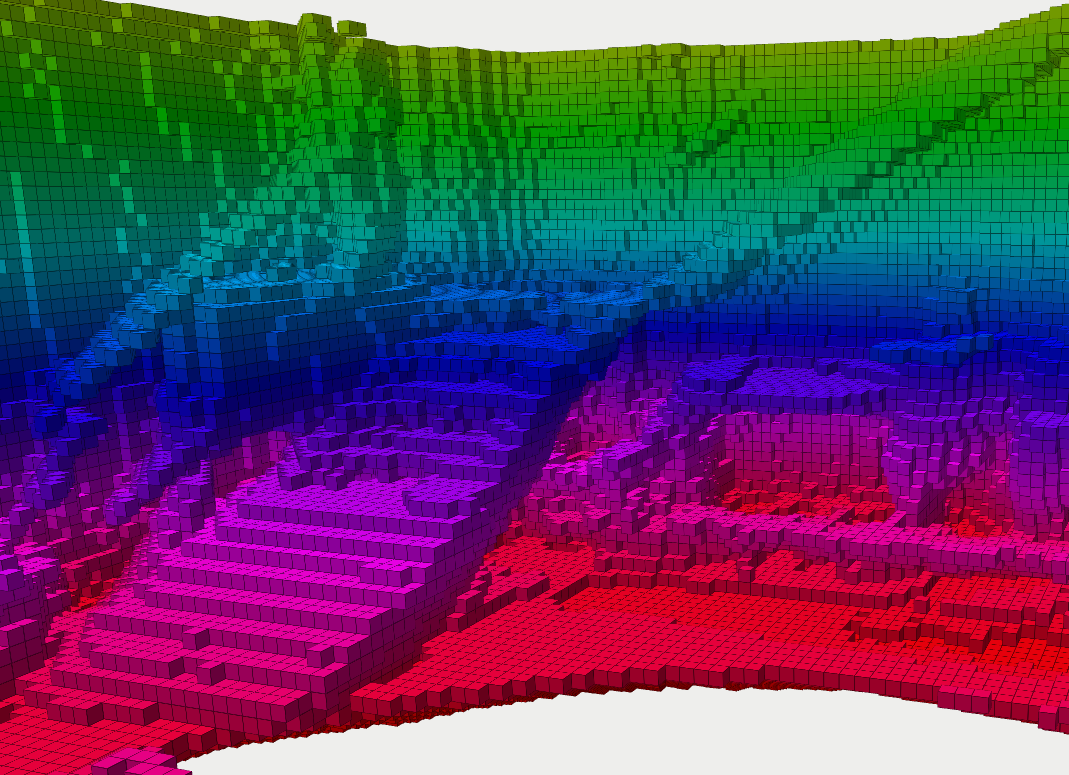}
    \includegraphics[width=0.239\textwidth]{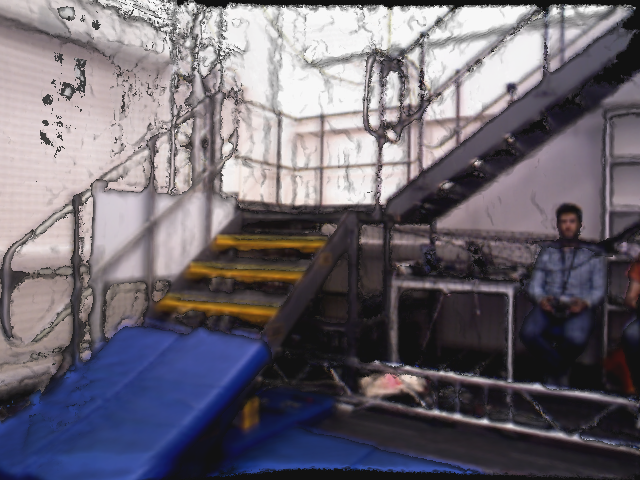}
    \caption{In this work, a light-weight exploration algorithm for MAVs was
        developed. [Left] The map created on-board the MAV in real-time using a
        voxel resolution of 10 cm which is used for exploration planning and
        collision avoidance. [Right] A high resolution render of a map created
        offline from the depth and colour images collected in the experiment.
        The map was created using the pipeline presented in
        \cite{Vespa_20193DV}.}
    \label{fig:cover}
    \vspace{-1ex}
\end{figure}

\section{Related Work}
A considerable amount of work is available on autonomous exploration. The
utility metrics that drive the exploration can be grouped into two categories,
map entropy metrics \cite{kumar2015cauchyschwarz} and unknown volume metrics
\cite{Witting2018historyaware}. These metrics can be utilised by either of the
two main exploration strategies which are sampling-based and frontier-based
exploration.

In the original work on frontier exploration \cite{yamauchi1997}, the closest
frontier from the current robot position is chosen. In \cite{cieslewski2017},
priority is given to maintaining a high MAV velocity during the exploration
task. Gao et al.\ \cite{Gao2018AnIF} focus on keeping track of visited key
nodes in a topological graph while conducting frontier exploration. A more
recent work on frontier-based autonomous inspection is presented in
\cite{Faria2019}.

The core idea of the sampling-based exploration approach is to sample potential
poses which could contribute to mapping the unknown volume. This strategy
avoids the map-wide computation of frontiers but requires the evaluation of
each sample. Bircher et al.\ \cite{Bircher_AR2018} have shown that the
Next-Best-View (NBV) method \cite{Connolly1985TheDO} can be applied in 3D
exploration. The NBV is found by growing an RRT to sample a position and yaw
angle in free space. This method is improved in \cite{Witting2018historyaware}
where a history of visited locations is maintained to avoid being stuck at
local minima.

Exploration strategies using map entropy as the evaluation metric  mostly use
the Shannon entropy \cite{Palazzolo2017,Scott2018MultirobotEA,Palazzolo2018}.
In \cite{Carillo2017}, the combination of the Shannon and the Renyi entropy
provides a utility function which balances between the robot localisation and
map uncertainty. In \cite{kaufman2018}, mapping is performed in 3D while motion
planning is done in 2D to reduce computational complexity since an MAV
typically flies at a constant height. Bissmarck et al.\ \cite{tolt2015}
compared various approaches to compute the information gain for candidate
views. The proposed frontier oriented volumetric hierarchical ray tracing was
benchmarked against the hierarchical ray-tracing by Vasquez-Gomez et al.\
\cite{VasquezGomez2013HierarchicalRT} in computation time, mapping efficiency
and estimation error.

The second group of evaluation metrics estimates the amount of unknown volume
in the view frustum via ray casting. This metric is used in
\cite{Witting2018historyaware,Bircher_AR2018,delmerico2018comparison,Kostas2018VisualSR,Kostas2017Uncertainty}.
The latter two add secondary tasks which are achieved in parallel to the main
exploration task.

\section{Problem Statement}
Let $V \subset \mathbb{R}^3$ be a bounded volume whose points $\mathbf{v} \in
V$ have the occupancy probabilities $P_o(\mathbf{v})$.  The goal of autonomous
exploration is to use a sensor-equipped robotic platform to create a map $M$ of
all the observable space in $V$.  In most non-trivial environments, there are
points $V_\mathrm{unob} \subset V$ that can not be observed by the sensor, such
as the interior of solid objects or passages too small for the robotic platform
to explore.  Thus the goal of autonomous exploration is to update the occupancy
probability of all observable points by the sensors $V_\mathrm{obs} = V
\setminus V_\mathrm{unob}$ to \textit{free} or \textit{occupied}.  Another
concept useful in exploration is that of frontiers.  Frontiers are the
boundaries between \textit{free} and \textit{unknown} space.  They mark
observable regions that can lead to an expansion of the map if observed.  The
exploration termination condition can be equivalently expressed in terms of
frontiers as the state where there are no more frontiers left.

Initially, all points $\mathbf{v} \in V$ are \textit{unknown} if no prior
information is available and the robotic platform has to map and navigate the
unknown space.  Path planning, collision avoidance, and exploration planning
have to be performed online as there is no \emph{a priori} knowledge of the
environment.  Thus it is important for the whole exploration pipeline to run
efficiently on-board the robotic platform.

In this paper we propose a solution to autonomous exploration using an MAV that
is equipped with a depth sensor and is employing a volumetric map to represent
the environment. The presented exploration strategy is iterative and is tightly
coupled with the underlying map representation, as detailed in the following.

\section{Frontier-based Information-driven Exploration Algorithm}
We propose a frontier sampling, information gain-based exploration approach.
It takes inspiration from both frontier-based and sampling-based exploration
methodologies.  Frontiers are good indicators of the regions the exploration
should be focused on, which are ignored by conventional sampling-based
exploration algorithms.  By sampling candidate next-views at the frontiers
there is no need for clustering of the individual voxels into larger frontiers,
thus removing the associated large computational cost.
A utility function expressing expected information gain over time is used to
evaluate the candidate next-views. This is a more meaningful metric than just
counting unobserved voxels and does not require a tuning parameter to take the
path cost into account.

\begin{figure}[htb]
    \centering
    \includegraphics[width=0.48\textwidth]{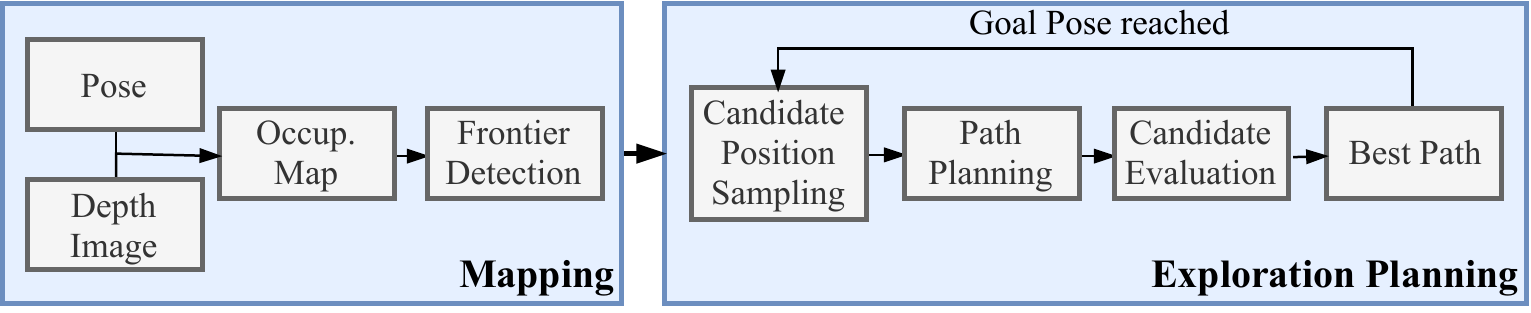}
    \caption{The mapping system runs continuously in the background while the exploration planner produces a new path each time the previous path is completed.}
    \label{fig:algorith_overview}
\end{figure}

An occupancy mapping server \cite{Vespa_RAL2018} is kept running in the
background, which continuously integrates depth image and pose pairs into the
map and extracts the frontiers. At the beginning of the exploration and after
some measurements have been integrated into the map, the exploration planner is
called.
At each planning iteration it executes the following steps in sequence:
\begin{enumerate}
	\item A pre-defined number of candidate goal positions are sampled from
		frontier voxels (Section \ref{sec:sampling}).
	\item Collision-free paths from the current position to each candidate
		position are planned (Section \ref{sec:path}).
	\item A yaw angle is associated to each candidate position, creating a
		candidate pose, and each pose is evaluated based on a utility function
		expressing information gain over time (Section \ref{sec:evaluation}).
	\item The candidate pose with the highest utility is selected as the next
		goal.
\end{enumerate}
The map is continuously updated as the MAV moves towards the goal. Once
the goal is reached, the next planning iteration begins. The exploration is
considered complete when there are no more frontier clusters remaining. A
graphical overview of the proposed approach is presented in Figure
\ref{fig:algorith_overview}.

\subsection{MAV Model}
Our exploration method is specifically tailored for MAVs. It is assumed that
the MAV's state $\mathbf{x}$ consists of its position $\mathbf{p} = [x, y, z]^T
\in V$ and yaw angle, $\psi \in [0, 2\pi)$, relative to the World frame
$\cframe{W}$ thus $\mathbf{x} = [x, y, z, \psi]^T \in V \times [0, 2\pi)$. It
is also assumed that the MAV has a maximum linear velocity $\upsilon_{\max} \in
\mathbb{R}^+$ and maximum yaw rate $\omega_{\max} \in \mathbb{R}^+$. The MAV is
also equipped with a depth camera mounted on a fixed position, with maximum
sensor range $d_{\max} \in \mathbb{R}^+$ and horizontal and vertical field of
view $\alpha_h$ and $\alpha_v$ respectively. For collision checking, we assume
the MAV is contained inside a sphere with centre $\mathbf{p}$ and safety radius
$R \in \mathbb{R}^+$.

\subsection{Map Representation}
We chose to use an occupancy map for three main reasons. First, it provides an
explicit representation of free space in which collision-free paths can be
easily planned. Second, an occupancy map is also inherently suitable for
integrating noisy sensor measurements. Finally, it provides an implicit measure
of map quality through its entropy.

The occupancy mapping framework used in this work is \textit{supereight}
\cite{Vespa_RAL2018}. \textit{supereight} uses an octree structure to store the
map. An octree is a tree data structure each node of which
subdivides the 3D space it corresponds to into 8 octants which are represented
as its children nodes in the octree. \textit{supereight}
also provides efficient map updates and queries by using Morton codes
for spatial indexing.  Instead of storing single voxels at the octree leaf
level, \textit{supereight} stores blocks of voxels at the leaf level, typically
$8 \times 8 \times 8$ voxels in size.  The side length in meters of a single
voxel is called map resolution and is denoted $r$.  The map is continuously
updated as new pose and depth image pairs arrive. The system does not assume
any particular pose tracking method. Thus any system can be run in parallel to
the exploration and provide an estimate of the MAV's pose, e.g.\ a
visual-inertial odometry pipeline or a motion capture system.

We extend \textit{supereight} to store whether each voxel is a
\textit{frontier} and perform frontier detection, as described in the following
section.  Additionally, up-propagation of occupancy probabilities through the
octree levels is performed to facilitate efficient collision queries: after a
new depth measurement is integrated into the map, each parent node whose
children were updated will also have its occupancy probability updated.
Specifically, the occupancy probability of the parent is set to the maximum of
the occupancy probabilities of its children. This way, if an octree node has an
occupancy probability smaller than 0.5, it is guaranteed that its children also
have an occupancy probability smaller than 0.5.

\subsection{Frontier Detection}
The map frontiers are detected during the map update process.  The frontiers
are stored as a sorted list $F$ containing Morton codes corresponding to voxel
blocks (octree leaves) with one or more frontier voxels. An implicit
frontier clustering at the voxel block level is achieved by considering voxels
in the same voxel block as a frontier cluster. Thus any computationally
intensive clustering methods as in \cite{kumar2015cauchyschwarz} are avoided.

At each depth measurement integration into the map, the voxels inside the
camera frustum are updated. Each of the updated voxels is tested for being a
frontier voxel. Voxels are considered frontiers if their occupancy probability
is lower than 0.5, while one or more of their 6 face neighbour voxels has an
occupancy probability of exactly 0.5. More intuitively, frontier voxels are
free voxels located next to completely unobserved voxels. The Morton code list
is updated as new frontier voxels emerge and previously unobserved regions are
observed. The latter leads to removal of frontier voxels. By only considering
the last updated voxels for the map frontiers update, a map-wide operation is
avoided. This frontier update is performed continuously at each sensor
measurement integration.

\subsection{Candidate Position Sampling}
\label{sec:sampling}
At the beginning of each planning iteration, a predefined number $N_c$ of
candidate positions $\mathbf{p}_i = [x_i, y_i, z_i]^T \in \mathbb{R}^3, \ i \in
\{1 \dots N_c\}$ is uniformly sampled from the frontier voxel blocks in $F$.
Initially, the number of frontier voxels in each frontier voxel block is
obtained. Frontier voxel blocks containing fewer frontier voxels given a
certain threshold are ignored for the duration of the candidate position
sampling, resulting in the filtered frontier list $\tilde{F}$. Since
$\tilde{F}$ is sorted and due to the spatial indexing of Morton codes, a more
spatially uniform candidate position sampling can be achieved by sampling every
$\lceil N_{rem}/N_c\rceil$th frontier voxel block in $\tilde{F}$, where
$N_{rem}$ is the number of frontier voxel blocks in $\tilde{F}$ and $\lceil
\cdot \rceil$ denotes the ceiling of the argument. For each of the selected
frontier voxel blocks, one of its frontier voxels is randomly selected and its
coordinates used as a candidate position $\mathbf{p}_i$. The current MAV
position is always added to the candidates to ensure that a rotation around the
yaw axis without any movement is also considered as candidate pose.

\subsection{Path Planning to Candidate Positions}
\label{sec:path}
The Open Motion Planning Library (OMPL) \cite{Sucan_2012IEEERAM} is used for
computing paths to the candidate positions. The integrated planning algorithm
is the informed RRT* \cite{Gammell2014InformedRO} with path simplification. The
octree map structure is exploited to perform efficient collision checking. Free
space queries are made by checking the occupancy probability against a
threshold and are first performed at a higher octree level, before checking at
the single-voxel level. A sphere around the MAV is used for collision checking
for points and a cylinder for line segments. The path is computed in
$\mathbb{R}^3$ and the corresponding yaw angles are computed at a later stage.

Since candidate positions are sampled at the frontier, hence close to unknown
space, the MAV cannot safely reach them due to its finite dimensions. In this
case, OMPL computes a collision-free path which ends before the candidate
position, typically at a distance near the MAV's safety radius. The candidate
position is then updated to this safe path endpoint resulting in the path
$W_i(\mathbf{p}, \mathbf{p}_i)$ from the current MAV position $\mathbf{p}$ to
the updated candidate. The path consists of one or more line segments.

In receding horizon exploration methods \cite{Bircher_AR2018,
Selin_2019IEEERAL}, the MAV moves only to the first point of a planned path
before performing another planning iteration. This sometimes has the effect
that the MAV moves back and forth in a small area since at subsequent planning
iterations it may move to the first point of paths to different endpoints
(i.e.\ from different frontiers). Thus the MAV is temporarily stuck in a small
region. In our approach, the MAV follows the path to its endpoint, committing
to exploring a single frontier before moving towards another goal.

\subsection{Yaw Optimisation and Candidate Pose Evaluation}
\label{sec:evaluation}
The map entropy is computed at each candidate position by performing a $360^o$
sparse raycasting. Each candidate position $\mathbf{p}_i$ is converted
into a candidate pose $\mathbf{x}_i = [x_i, y_i, z_i, \psi_i]^T$ by optimising
the yaw angle for the highest entropy given the $360^o$ raycast. The estimated
travel time to the candidate pose is computed based on the MAV's maximum linear
velocity $\upsilon_{\max}$ and maximum yaw rate $\omega_{\max}$.  The utility
function of a candidate goal pose is computed as the map entropy over the
estimated path travel time. Finally the candidate pose with the highest utility
is selected as the next goal pose.
We will detail these steps in the following.

\subsubsection{Sparse Raycasting and Yaw Optimisation}
Instead of random sampling the MAV yaw angle using the informed RRT*, it is
optimised by performing a sparse $360^o$ raycasting which at the same time
computes the map entropy along each ray.  It has been shown
\cite{Oleynikova_2018IEEERAL} that sparse raycasting can effectively
approximate a full raycasting at a fraction of the computational cost.

Given the sensor's vertical field of view $\alpha_v$ and its maximum range
$d_{\max}$, the raycasting is performed with a horizontal angle increment
$\delta \psi$ for the yaw angle and a vertical angle increment $\delta \theta$
for $\alpha_v$, with the rays being stopped at a distance $d_{\max}$ from
$\mathbf{p}_i$.  The entropy for a single voxel $\mathbf{v}$ is calculated
using Shannon's information theory
\begin{equation}
    \mathbb{H}(\mathbf{v}) = -P_o(\mathbf{v}) \ln P_o(\mathbf{v}) - \left( 1 - P_o(\mathbf{v}) \right) \ln \left( 1 - P_o(\mathbf{v}) \right),
\end{equation}
where $\mathbb{H}(\mathbf{v})$ denotes the map entropy and $P_o(\mathbf{v})$
the occupancy probability of a voxel $\mathbf{v}$.

\begin{figure}[htb]
    \vspace{1ex}
    \centering
    \includegraphics[width=0.30\textwidth]{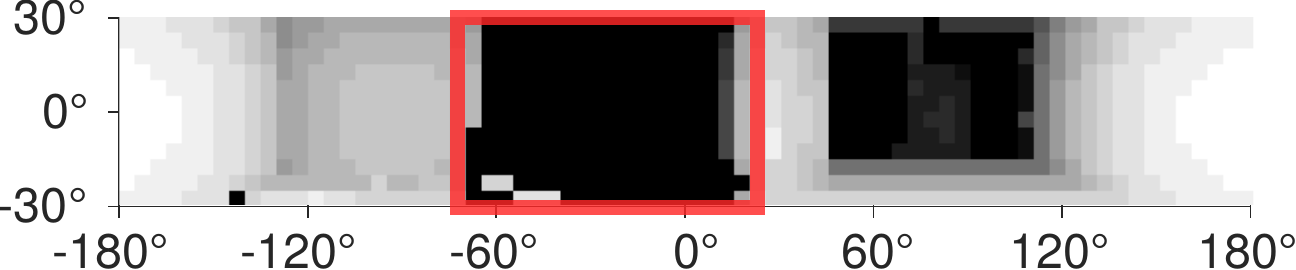} \\
    \vspace{1ex}
    \includegraphics[width=0.30\textwidth]{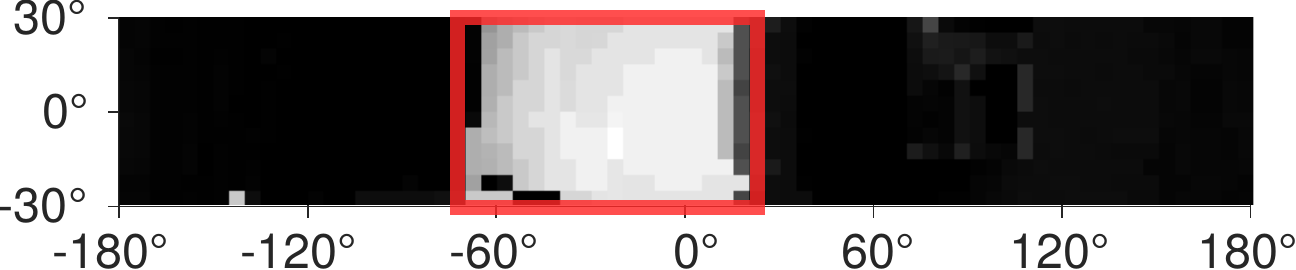}\\
    \vspace{-1ex}
    \caption{Sparse raycasted depth map [Top] and corresponding entropy
	map [Bottom]. The boundaries of the camera frustum at the yaw angle that
results in the highest map entropy are shown in red.}
    \label{fig:raycasting}
    \vspace{-1ex}
\end{figure}

The entropy along a ray is computed as the sum of the entropy of the voxels
along the ray until the first voxel $\mathbf{v}$ with $P_o(\mathbf{v}) > 0.5$
or $d_{\max}$ is reached, whichever comes first.
By taking into account the entropy of both observed and unobserved voxels, we
evaluate both unknown regions and regions where the map is still uncertain and
of low quality. Thus it acts as a measure of both map coverage and quality.
The raycasting is performed
on the current map and computes a $360^o$ estimated entropy map, with each
pixel corresponding to a single ray, like the one shown in Figure
\ref{fig:raycasting} [Bottom].  The brighter a pixel is, the higher the entropy
for the corresponding ray.  The corresponding sparse depth map is shown in
Figure \ref{fig:raycasting} [Top].  Lighter shades of grey denote rays that hit
an occupied voxel closer to the candidate position, while black denotes rays
that did not hit any occupied voxels up to a distance $d_{\max}$, producing an
invalid depth measurement.
This can be either due to free space extending to a distance greater than
$d_{\max}$ or due to the existence of voxels with high map entropy, as is the
case for the rightmost and leftmost invalid depth regions respectively in
Figure \ref{fig:raycasting} [Top].

In order to obtain the optimal yaw for the candidate position, a sliding window
summation with an angular width $\alpha_h$ is performed on the entropy map.
For each yaw value, the sum of the entropy of the rays that fall inside the
sensor's field of view is computed. The yaw angle $\psi_i$ with the highest
cumulative map entropy is selected, resulting in the candidate pose
$\mathbf{x}_i$. We use $\mathbb{H}(\mathbf{x}_i)$ to denote the map entropy
obtained by raycasting from candidate pose $\mathbf{x}_i$. The boundaries of
the camera frustum at the yaw angle that results in the highest map entropy are
shown in red in Figure \ref{fig:raycasting}.

\subsubsection{Utility Function} The utility for a candidate pose
$\mathbf{x}_i$ given the path $W_i$ to it is the entropy of the portion of the
map visible from the pose over the time it takes to complete the path $W_i$.
This results in a balance between exploring the MAV's immediate surroundings by
penalizing time-consuming paths and moving towards more distant frontiers
leading to potentially completely unexplored regions by favouring high map
entropy. By formulating the utility as a ratio, this balance is achieved
without the need for any tuning parameter.

The estimated information gain computation using sparse raycasting was
described in the previous section.  The estimated time $T(W_i)$ to complete
path $W_i$ is computed by assuming the MAV always flies at its maximum linear
speed and rotates at the maximum rate when needed.  Thus the total path travel
time $T(W_i)$ is estimated as the maximum between the time the MAV needs to
move from its start point $\mathbf{p}$ to its final point $\mathbf{p}_i$ at
$\upsilon_{\max}$ and the time it takes to rotate from the start yaw $\psi$ to
its final yaw $\psi_i$ at $\omega_{\max}$.  Although this estimate is not
accurate due to the constraints imposed by the MAV's dynamics, it is a suitable
approximation. The utility function serves to provide a ranking of the
candidate poses.

Thus the utility of a candidate pose is computed as
\begin{equation}
    u(\mathbf{x}_i, \hat{W_i}) = \frac{\mathbb{H}(\mathbf{x}_i)}{T(\hat{W_i})},
\end{equation}
and the candidate pose $\hat{\mathbf{x}}_i = [\hat{\mathbf{p}}_i^T
\hat{\psi}_i]^T$ with the highest utility is selected as the next goal pose.

Since the path $W_i(\mathbf{p}, \hat{\mathbf{p}}_i)$ consists of a number of
points with no associated yaw, each point in the path has to be matched to a
yaw angle before it is supplied to the MAV's controller. The initial and final
points of the path already have associated yaw angles $\psi$ and $\hat{\psi}_i$
from the current $\mathbf{x}$ and best candidate pose $\hat{\mathbf{x}}_i$
respectively. For each intermediate path point a yaw angle is assigned by
performing the same yaw optimisation as for $W_i(\mathbf{p}, \mathbf{p}_i)$,
resulting in the final path $\hat{W_i}(\mathbf{x}, \hat{\mathbf{x}}_i)$ to the
goal pose. Essentially, for each intermediate point of the path to the best
candidate pose a sparse raycasting and related map entropy computation is
performed.
The number of intermediate points depends on the geometry of the space but due
to the OMPL path simplification it is in general low. Since this intermediate
yaw optimisation is performed only for the path to the best candidate, its
effect on the total computation time is limited.

\section{Experimental Evaluation}
The exploration algorithm has been evaluated in both simulated and real world
experiments.  All simulated experiments were performed in Ubuntu 18.04 using
ROS Melodic and the RotorS simulator \cite{Furrer2016Rotors}. The MAV model
used in the simulation is the Ascending Technologies FireFly hexacopter
equipped with a VI-Sensor mounted at a $15^{\circ}$ angle. For all simulations
20 candidate poses are sampled at each planner iteration. All simulations were
run on an Intel Core i7-8750H CPU operating at 2.20 GHz and compiled with g++
7.4.0 using the O3 optimization level.
The parameters used for both the simulated and real-world experiments can be
found in Table \ref{table:params}.

\subsection{Apartment Environment Simulation}
\begin{figure}[htb]
    \vspace{1ex}
    \centering
    \includegraphics[width=0.37\textwidth]{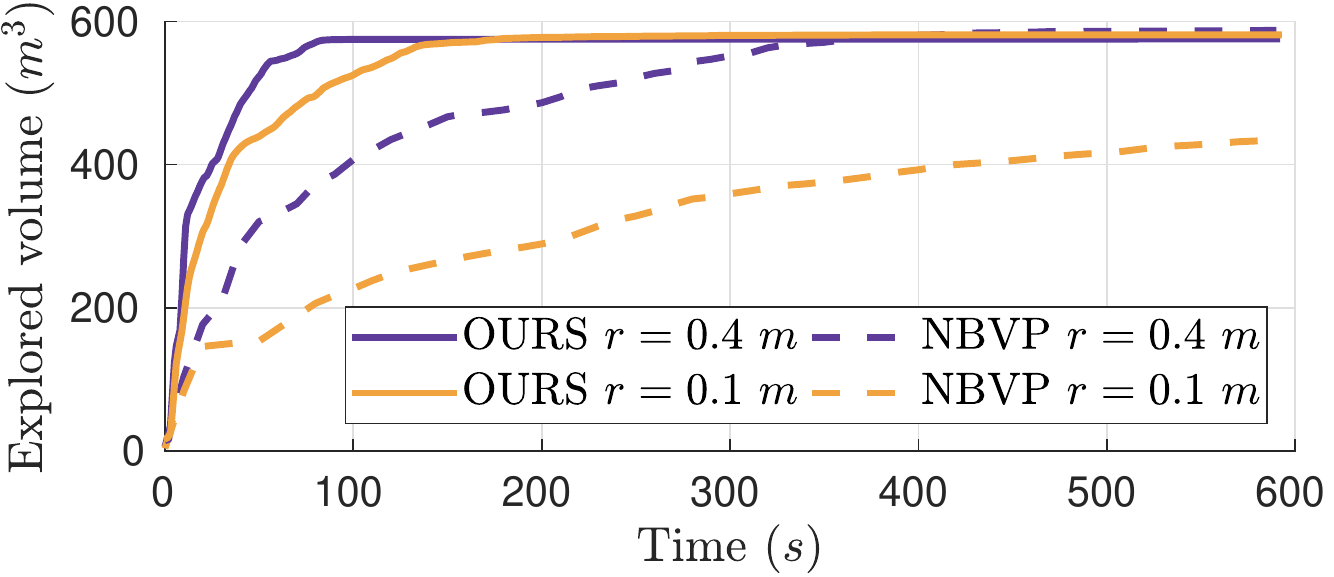}
    \vspace{-1ex}
    \caption{Apartment environment explored volume over time.}
    \label{fig:apartment_plot}
    \vspace{-1ex}
\end{figure}

Several simulations were conducted in the 10 m $\times$ 20 m  $\times$ 3 m
apartment environment used in \cite{Bircher_AR2018,Selin_2019IEEERAL}. Our
algorithm was compared with the NBV planner presented in \cite{Bircher_AR2018}.
Figure \ref{fig:apartment_plot} shows the explored volume over time, averaged
over 10 runs, for both algorithms using a voxel resolution of $r = 0.1$ m and
$r = 0.4$ m. It can be observed that our method explores the whole environment
substantially faster, especially when using a high resolution map. An
interesting property of our method is that it does not require any arbitrary
initialisation movement, like the $360^{\circ}$ in-place rotation in
\cite{Bircher_AR2018}. Instead, the utility function usually gives higher
utility to candidate poses that result in the MAV performing a similar rotation
in the beginning of the exploration. Our method explores $95\%$ of the
environment in 80 s and 151 s for a resolution of 0.4 m and 0.1 m respectively.
These are lower than the respective times reported in \cite{Selin_2019IEEERAL},
although it should be noted that we were unable to run their planner on our
hardware, despite our best efforts, so the results might not be directly
comparable, especially since the hardware used by the authors is not stated.

\begin{table}[htb]
    \centering
    \caption{Exploration parameters}
    \label{table:params}
    \begin{tabular}{ | l | c c c c | }
        \hline
        Parameter               & Apartment                  & Maze                        & Powerplant                  & Experiment \\ \hline
        $r$ (m)                 & 0.1, 0.4                   & 0.1, 0.2                    & 0.2                         & 0.1 \\
        $R$ (m)                 & 0.5                        & 0.5                         & 0.5                         & 0.8 \\
        $\upsilon_{\max}$ (m/s) & 1.5                        & 1.5                         & 0.7, 1.5, 2.5               & 0.1 \\
        $\omega_{\max}$ (rad/s) & 0.75                       & 0.75                        & 0.75                        & 0.15 \\
        $d_{\max}$ (m)          & 5                          & 5                           & 7                           & 4 \\
        $[\alpha_h, \alpha_v]$  & $[90^{\circ}, 60^{\circ}]$ & $[115^{\circ}, 60^{\circ}]$ & $[115^{\circ}, 60^{\circ}]$ & $[58^{\circ}, 45^{\circ}]$ \\ \hline
    \end{tabular}
    \vspace{-1ex}
\end{table}

\subsection{Maze Environment Simulation}
\begin{figure}[htb]
    \centering
    \includegraphics[width=0.37\textwidth]{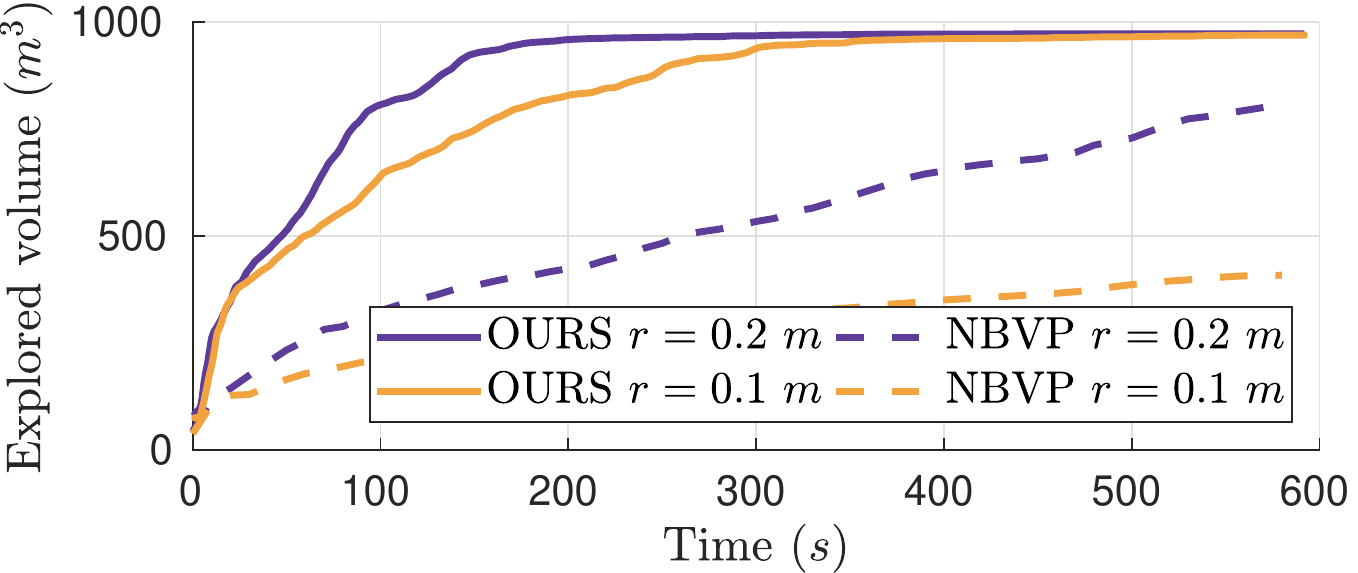}
    \caption{Maze environment explored volume over time.}
    \label{fig:maze_plot}
    \vspace{-1ex}
\end{figure}

\begin{figure}[htb]
    \centering
    \includegraphics[width=0.23\textwidth]{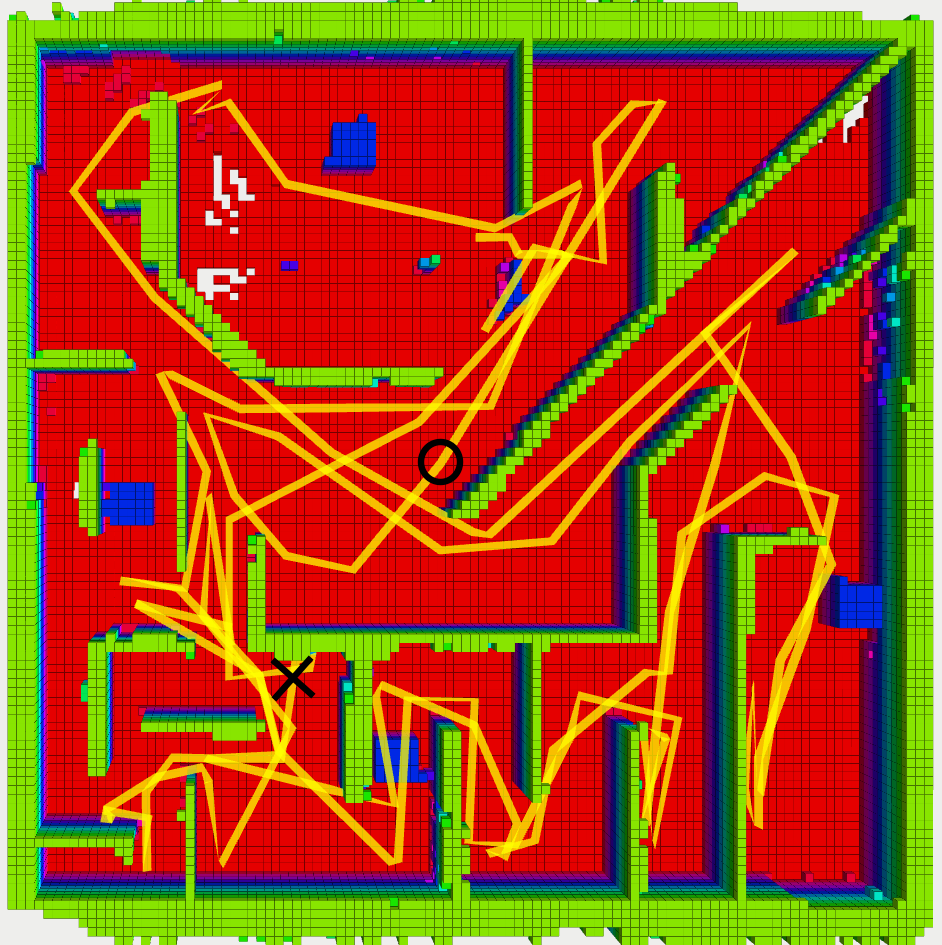}
	\caption{Top-down view of a map of the maze environment created by our
	algorithm.  Voxels are coloured based on their height, with red indicating
	floor voxels and green voxels near the cropped roof. The MAV's starting
	position is marked by a black circle, it's path by a yellow line and it's
	final position by a black cross.}
    \label{fig:maze_path}
    \vspace{-1ex}
\end{figure}

Simulations were also conducted in a more complex 20 m $\times$ 20 m $\times$
2.5 m maze environment from \cite{Oleynikova_2017IROS}. Figure
\ref{fig:maze_plot} shows the explored volume over time, averaged over 10 runs,
for both our algorithm and NBVP using a voxel resolution of 0.1 m and 0.2 m. The
planners were not run at a resolution of 0.4 m because this environment
contains some narrow corridors that the MAV cannot navigate through when such a
coarse resolution is used. Our method performs significantly faster than NBVP in
this case since the random sampling of NBVP does not always detect unexplored
regions quickly. This results in the MAV being stuck in a portion of the map
for a long time before moving towards unexplored regions. Our planner does not
exhibit this behaviour due to its use of frontiers which reliably guide the MAV
towards unexplored space. Our method explores $95\%$ of the environment in 177
s and 330 s for a resolution of 0.2 m and 0.1 m respectively. Figure
\ref{fig:maze_path} shows a top-down view of a map created by our algorithm in
the maze environment.

\subsection{Powerplant Environment Simulation}
\begin{figure}[htb]
    \centering
    \includegraphics[width=0.21\textwidth]{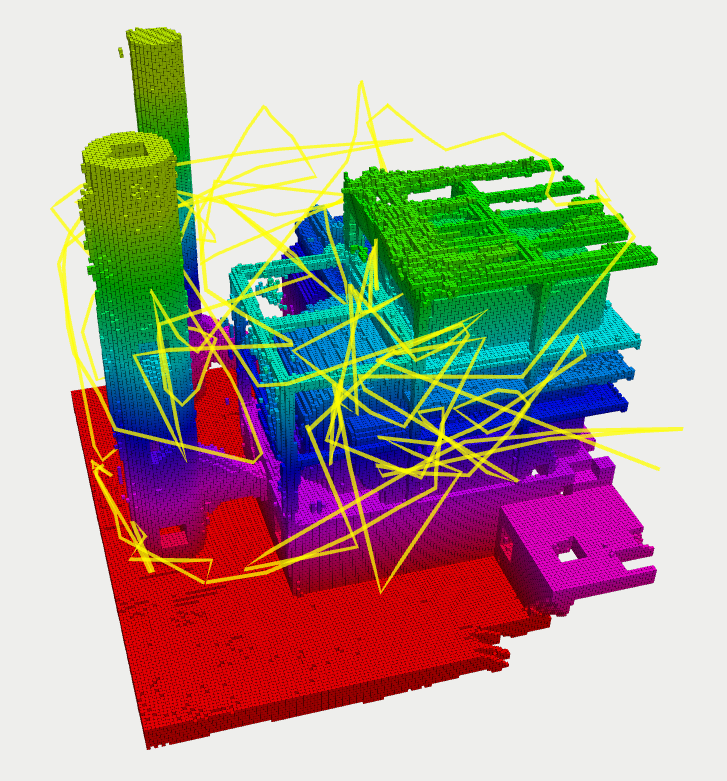}
    \includegraphics[width=0.26\textwidth]{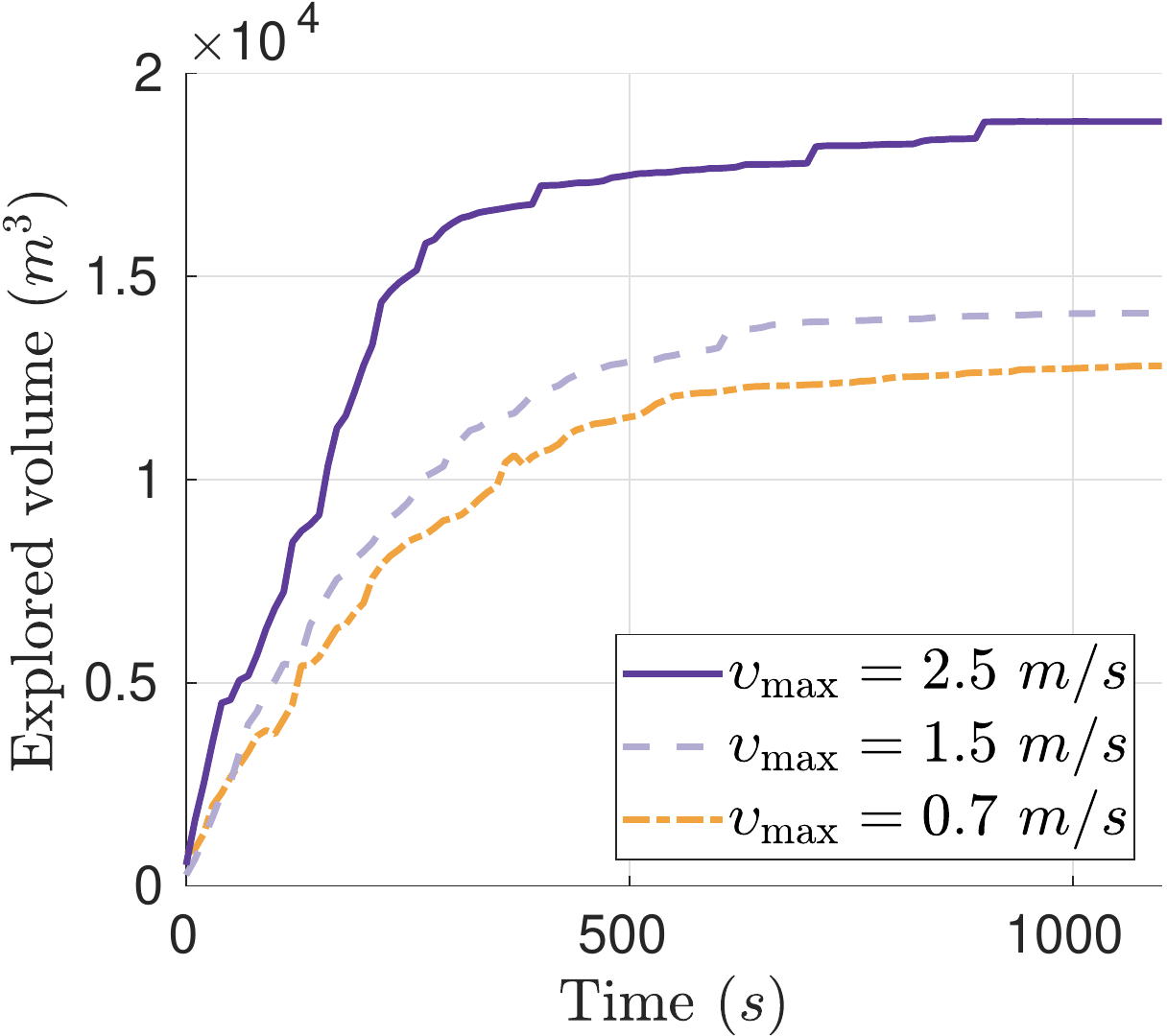}
	\caption{[Left] A map of the powerplant environment created by our method.
	Voxels are coloured based on their height and the MAV's path through the
	environment is marked by a yellow line. [Right] Powerplant environment
	explored volume over time.}
    \label{fig:powerplant_path}
    \vspace{-1ex}
\end{figure}

In order to evaluate our method in an outdoors scenario, the 33 m $\times$ 31 m
$\times$ 26 m powerplant environment from the Gazebo simulator was used.
Figure \ref{fig:powerplant_path} [Right] shows the explored volume over time,
averaged over 10 runs, for various MAV maximum linear velocities. Figure
\ref{fig:powerplant_path} [Left] shows a map created by our algorithm and the
corresponding MAV path. In this simulation, the planner is unable to explore
the whole space because a large portion of it is unobservable, e.g.\ the
interior of the building. For lower MAV velocities the exploration is slower
not only due to the lower movement speed but because candidate views further
away have a lower information gain over time due to the increased time it takes
to fly to then. Thus the MAV more thoroughly explores the nearby regions before
moving to ones further away. The same environment has been used to evaluate
\cite{Cieslewski_2017IROS} and \cite{Selin_2019IEEERAL} by reporting only the
time required to complete the exploration. However, since both
\cite{Cieslewski_2017IROS} and \cite{Selin_2019IEEERAL} use different
exploration termination conditions, a comparison of the completion time would
not necessarily be indicative of a planner's performance. A more meaningful
comparison could be made by comparing the amount of volume explored in a given
amount of time, as presented in Figure \ref{fig:powerplant_path} [Right].

\subsection{Computational Complexity}
Since the planner is running on-board an MAV its computational complexity is an
important characteristic. Sampling the candidate positions has a complexity of
$\bigO(N_c)$ and computing a single collision free path with $N_p$ points is
$\bigO(\frac{N_p - 1}{r^3})$. Performing a single sparse raycast with $n$
horizontal rays and $m$ vertical rays scales with $\bigO(\frac{n m}{r})$ while
the yaw optimization scales with $\bigO(n)$. Computing the utility of a single
path requires computing its length, thus it scales with $\bigO(N_p)$. Finally,
for the best path only, the yaw at the intermediate path points is computed
with a sparse raycast and yaw optimization for each point giving $\bigO(N_p
(\frac{n m}{r} + n))$. Putting it all together, the computational complexity
for a single planner iteration is
$\bigO \left(
    N_c \left( \frac{N_p}{r^3} + \frac{n m}{r} + n + N_p \right)
    + N_p \left( \frac{n m}{r} + n \right)
\right)$.

For a quantitative comparison to other methods the per-iteration planner
computation time for all simulations was measured. The mean and standard
deviation for the three simulated environments at different map resolutions for
both our planner and NBVP are presented in Table \ref{table:timings}. The
reduction in computational time at higher resolutions can be attributed to the
sparse raycasting, the efficient collision checking performed by our method.
It should be noted that receding horizon planners like \cite{Bircher_AR2018}
and \cite{Selin_2019IEEERAL} typically perform a large number of planning
iterations since they perform only a portion of the computed path at each
iteration. Our algorithm performs relatively few planning iterations since each
computed path is followed until its end before replanning.

\begin{table}[htb]
    \centering
    \caption{Per-iteration computation time}
    \label{table:timings}
    \begin{tabular}{ | c c | r@{\hspace{0.3em}} l r@{\hspace{0.3em}} l r@{\hspace{0.4em}} l | }
        \hline
        & $r$ (m)
        & \multicolumn{2}{c}{Apartment (ms)}
        & \multicolumn{2}{c}{Maze (ms)}
        & \multicolumn{2}{c|}{Powerplant (ms)} \\ \hline
        \multirow{3}{*}{Ours}
        & 0.4  &  122 & $\pm$ 36   &      & \hspace{0.12em}--  &     & \hspace{0.12em}-- \\
        & 0.2  &  156 & $\pm$ 109  &  155 & $\pm$ 71           & 152 & $\pm$ 20          \\
        & 0.1  &   68 & $\pm$ 27   &  238 & $\pm$ 80           &     & \hspace{0.12em}-- \\
        \hline
        \multirow{3}{*}{NBVP}
        & 0.4  &   73 & $\pm$ 8    &      & \hspace{0.12em}--  &     & \hspace{0.12em}-- \\
        & 0.2  &  707 & $\pm$ 44   &  775 & $\pm$ 50           &     & \hspace{0.12em}-- \\
        & 0.1  & 7940 & $\pm$ 410  & 8540 & $\pm$ 425          &     & \hspace{0.12em}-- \\
        \hline
    \end{tabular}
    \vspace{-1ex}
\end{table}

\subsection{Real World Experiments}
\begin{figure}[htb]
    \centering
    \includegraphics[width=0.22\textwidth]{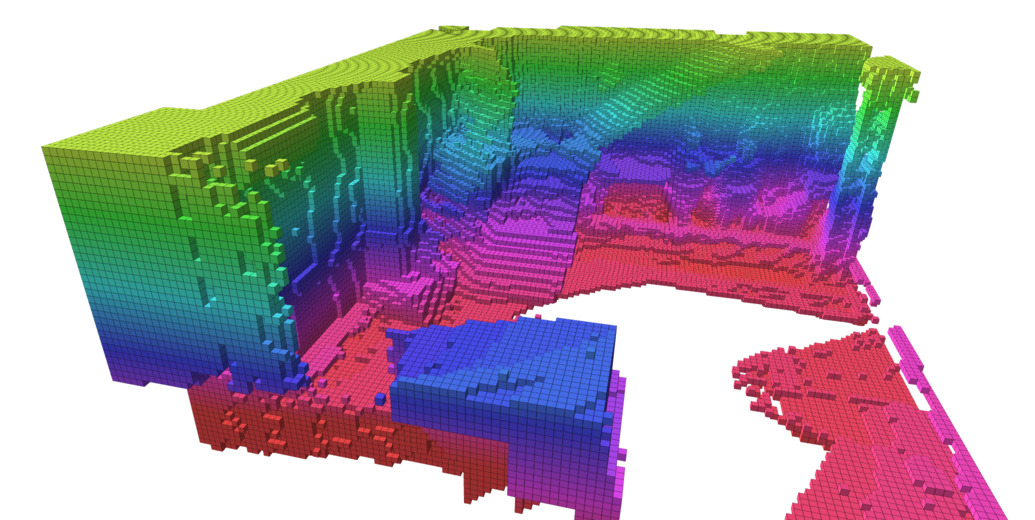}
    \includegraphics[width=0.2\textwidth]{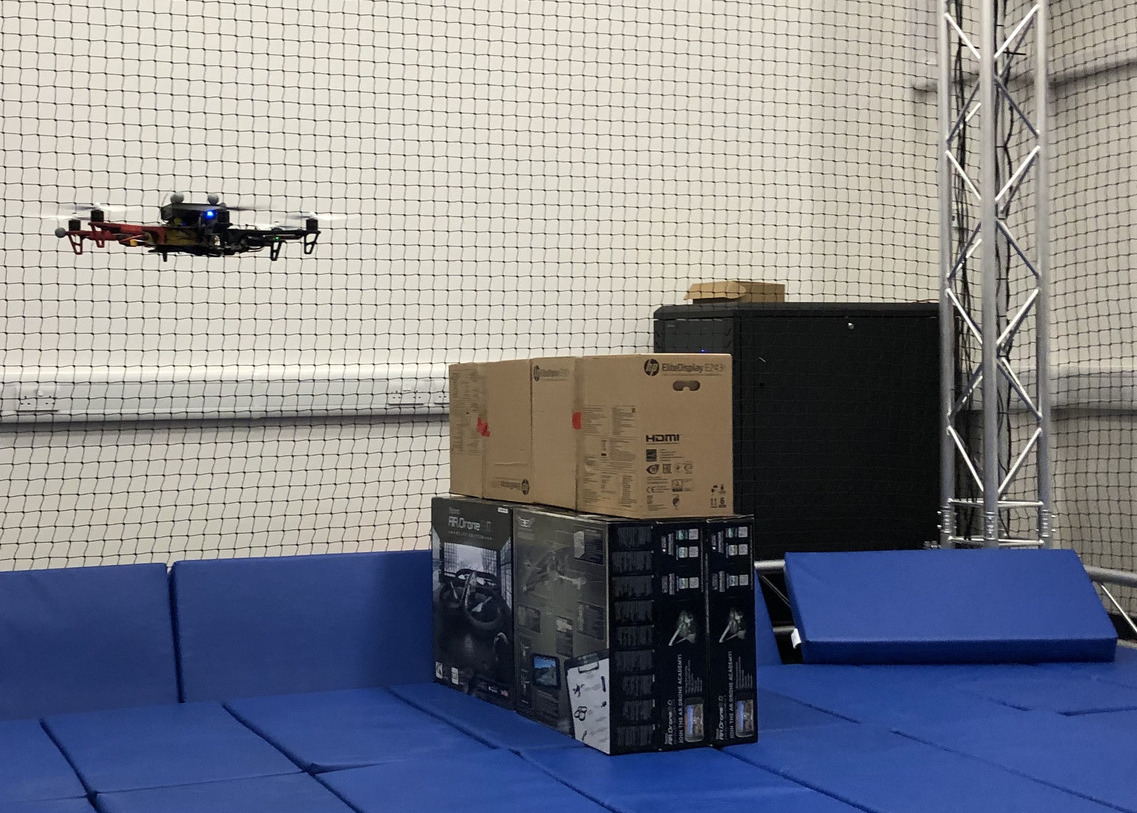}
	\caption{[Left] Map of the room the experiment was conducted in. Voxels are
	coloured based on their height and two of the walls have been cropped for
	visualization. [Right] A photograph of the MAV and the obstacle.}
    \label{fig:experiment}
    \vspace{-1ex}
\end{figure}

A real world experiment was performed in order to demonstrate the feasibility
of running the proposed exploration algorithm on-board an MAV. The experiment
was conducted in a 7 m $\times$ 5.5 m $\times$ 5 m room equipped with a VICON
motion capture system providing the MAV pose. The MAV used was a DJI F550
hexacopter equipped with an ASUS Xtion Pro RGBD camera. The entire system, both
mapping and exploration planning was run on-board the MAV on an Intel NUC with
an Intel Core i7-7567U CPU operating at 3.5 GHz. After a manual take-off to a
height of 1 m by the safety pilot, the exploration algorithm is initiated. A
0.5 m $\times$ 1.3 m $\times$ 0.8 m obstacle was placed inside the room in
order to create additional frontiers and force the MAV to navigate around it.
The mean per-iteration computation time was 383 ms with a standard deviation of
144 ms.  The resulting map can be seen in Figure \ref{fig:experiment} [Left]
and a photograph of the experimental setup in Figure \ref{fig:experiment}
[Right].

\section{Conclusions and Future Work}
In this work, we have presented an exploration strategy that is a hybrid
between frontier-based and sampling-based strategies. It improves upon the
state-of-the-art both in terms of exploration speed and computational cost as
shown in the simulation studies. The real-world experiment showcases that it is
feasible to use the proposed strategy in the real world and that it can be run
in real-time on-board an MAV.

In future work we are planning on using a multi-resolution mapping pipeline
like the one in \cite{Vespa_20193DV} and generating dynamically feasible drone
trajectories instead of a series of waypoints. More planned improvements
include the use of additional sensor types such as LIDARs, robust operation in
real-world outdoor environments and the ability to explore and navigate dynamic
environments.

%%%%%%%%%%%%%%%%%%%%%%%%%%%%%%%%%%%%%%%%%%%%%%%%%%%%%%%%%%%%%%%%%%%%%%%%%%%%%%%
% Bibliography
%%%%%%%%%%%%%%%%%%%%%%%%%%%%%%%%%%%%%%%%%%%%%%%%%%%%%%%%%%%%%%%%%%%%%%%%%%%%%%%
% Use to equalize the columns on the last page.
\IEEEtriggeratref{16}
\bibliographystyle{IEEEtran}
\bibliography{Bibliography/SP_bibliography,Bibliography/references}

\begin{thebibliography}{10}
\providecommand{\url}[1]{#1}
\csname url@samestyle\endcsname
\providecommand{\newblock}{\relax}
\providecommand{\bibinfo}[2]{#2}
\providecommand{\BIBentrySTDinterwordspacing}{\spaceskip=0pt\relax}
\providecommand{\BIBentryALTinterwordstretchfactor}{4}
\providecommand{\BIBentryALTinterwordspacing}{\spaceskip=\fontdimen2\font plus
\BIBentryALTinterwordstretchfactor\fontdimen3\font minus
  \fontdimen4\font\relax}
\providecommand{\BIBforeignlanguage}[2]{{%
\expandafter\ifx\csname l@#1\endcsname\relax
\typeout{** WARNING: IEEEtran.bst: No hyphenation pattern has been}%
\typeout{** loaded for the language `#1'. Using the pattern for}%
\typeout{** the default language instead.}%
\else
\language=\csname l@#1\endcsname
\fi
#2}}
\providecommand{\BIBdecl}{\relax}
\BIBdecl

\bibitem{Julia2012comparison}
M.~Juli{\'a}, A.~Gil, and {\'O}.~Reinoso, ``A comparison of path planning
  strategies for autonomous exploration and mapping of unknown environments,''
  \emph{Autonomous Robots}, vol.~33, pp. 427--444, 2012.

\bibitem{LaValle_PlanningAlgorithms2006}
S.~M. La{V}alle, \emph{Planning Algorithms}.\hskip 1em plus 0.5em minus
  0.4em\relax Cambridge University Press, 2006.

\bibitem{Vespa_20193DV}
E.~Vespa, N.~Funk, P.~H. Kelly, and S.~Leutenegger, ``Adaptive-resolution
  octree-based volumetric {SLAM},'' in \emph{2019 International Conference on
  3D Vision (3DV)}.\hskip 1em plus 0.5em minus 0.4em\relax IEEE, Sep 2019, pp.
  654--662.

\bibitem{kumar2015cauchyschwarz}
B.~{Charrow}, S.~{Liu}, V.~{Kumar}, and N.~{Michael}, ``Information-theoretic
  mapping using {C}auchy-{S}chwarz quadratic mutual information,'' in
  \emph{2015 IEEE International Conference on Robotics and Automation (ICRA)},
  May 2015, pp. 4791--4798.

\bibitem{Witting2018historyaware}
C.~{Witting}, M.~{Fehr}, R.~{Bähnemann}, H.~{Oleynikova}, and R.~{Siegwart},
  ``History-aware autonomous exploration in confined environments using
  {MAV}s,'' in \emph{2018 IEEE/RSJ International Conference on Intelligent
  Robots and Systems (IROS)}, Oct 2018, pp. 1--9.

\bibitem{yamauchi1997}
B.~Yamauchi, ``A frontier-based approach for autonomous exploration,'' in
  \emph{Proceedings of IEEE International Symposium on Computational
  Intelligence in Robotics and Automation, CIRA}, Oct 1997, pp. 146--151.

\bibitem{cieslewski2017}
T.~{Cieslewski}, E.~{Kaufmann}, and D.~{Scaramuzza}, ``Rapid exploration with
  multi-rotors: A frontier selection method for high speed flight,'' in
  \emph{2017 IEEE/RSJ International Conference on Intelligent Robots and
  Systems (IROS)}, Sep 2017, pp. 2135--2142.

\bibitem{Gao2018AnIF}
W.~Gao, M.~Booker, A.~H. Adiwahono, M.~Yuan, J.~Wang, and W.-Y. Yau, ``An
  improved frontier-based approach for autonomous exploration,'' \emph{2018
  15th International Conference on Control, Automation, Robotics and Vision
  (ICARCV)}, pp. 292--297, 2018.

\bibitem{Faria2019}
M.~Faria, I.~Maza, and A.~Viguria, ``Applying frontier cells based exploration
  and lazy theta* path planning over single grid-based world representation for
  autonomous inspection of large 3{D} structures with an {UAS},'' \emph{Journal
  of Intelligent {\&} Robotic Systems}, vol.~93, no.~1, pp. 113--133, Feb 2019.

\bibitem{Bircher_AR2018}
A.~Bircher, M.~Kamel, K.~Alexis, H.~Oleynikova, and R.~Siegwart, ``Receding
  horizon path planning for 3{D} exploration and surface inspection,''
  \emph{Autonomous Robots}, vol.~42, no.~2, pp. 291--306, Feb. 2018.

\bibitem{Connolly1985TheDO}
C.~I. Connolly, ``The determination of next best views,'' \emph{Proceedings.
  1985 IEEE International Conference on Robotics and Automation}, vol.~2, pp.
  432--435, 1985.

\bibitem{Palazzolo2017}
E.~Palazzolo and C.~Stachniss, ``Information-driven autonomous exploration for
  a vision-based {MAV},'' \emph{ISPRS Annals of Photogrammetry, Remote Sensing
  and Spatial Information Sciences}, vol. IV-2/W3, pp. 59--66, 08 2017.

\bibitem{Scott2018MultirobotEA}
M.~Scott and K.~Jerath, ``Multi-robot exploration and coverage: Entropy-based
  adaptive maps with adjacency control laws,'' \emph{2018 Annual American
  Control Conference (ACC)}, pp. 4403--4408, 2018.

\bibitem{Palazzolo2018}
E.~Palazzolo and C.~Stachniss, ``Effective exploration for {MAV}s based on the
  expected information gain,'' \emph{Drones}, vol.~2, no.~1, 2018.

\bibitem{Carillo2017}
H.~Carrillo, P.~Dames, V.~Kumar, and J.~Castellanos, ``Autonomous robotic
  exploration using a utility function based on {R}enyi's general theory of
  entropy,'' \emph{Autonomous Robots}, vol.~42, 08 2017.

\bibitem{kaufman2018}
E.~{Kaufman}, K.~{Takami}, Z.~{Ai}, and T.~{Lee}, ``Autonomous quadrotor 3{D}
  mapping and exploration using exact occupancy probabilities,'' in \emph{2018
  Second IEEE International Conference on Robotic Computing (IRC)}, Jan 2018,
  pp. 49--55.

\bibitem{tolt2015}
F.~{Bissmarck}, M.~{Svensson}, and G.~{Tolt}, ``Efficient algorithms for next
  best view evaluation,'' in \emph{2015 IEEE/RSJ International Conference on
  Intelligent Robots and Systems (IROS)}, Sep 2015, pp. 5876--5883.

\bibitem{VasquezGomez2013HierarchicalRT}
J.~I. Vasquez-Gomez, L.~E. Sucar, and R.~Murrieta-Cid, ``Hierarchical ray
  tracing for fast volumetric next-best-view planning,'' \emph{2013
  International Conference on Computer and Robot Vision}, pp. 181--187, 2013.

\bibitem{delmerico2018comparison}
J.~Delmerico, S.~Isler, R.~Sabzevari, and D.~Scaramuzza, ``A comparison of
  volumetric information gain metrics for active 3{D} object reconstruction,''
  \emph{Autonomous Robots}, vol.~42, no.~2, pp. 197--208, 2018.

\bibitem{Kostas2018VisualSR}
T.~Dang, C.~Papachristos, and K.~Alexis, ``Visual saliency-aware receding
  horizon autonomous exploration with application to aerial robotics,''
  \emph{2018 IEEE International Conference on Robotics and Automation (ICRA)},
  pp. 2526--2533, 2018.

\bibitem{Kostas2017Uncertainty}
C.~Papachristos, S.~Khattak, and K.~Alexis, ``Uncertainty-aware receding
  horizon exploration and mapping using aerial robots,'' in \emph{2017 IEEE
  international conference on robotics and automation (ICRA)}, 05 2017, pp.
  4568--4575.

\bibitem{Vespa_RAL2018}
E.~Vespa, N.~Nikolov, M.~Grimm, L.~Nardi, P.~H.~J. Kelly, and S.~Leutenegger,
  ``Efficient octree-based volumetric {SLAM} supporting signed-distance and
  occupancy mapping,'' \emph{IEEE Robotics and Automation Letters}, vol.~3,
  no.~2, pp. 1144--1151, Apr. 2018.

\bibitem{Sucan_2012IEEERAM}
I.~A. {Sucan}, M.~{Moll}, and L.~E. {Kavraki}, ``The open motion planning
  library,'' \emph{IEEE Robotics Automation Magazine}, vol.~19, no.~4, pp.
  72--82, Dec 2012.

\bibitem{Gammell2014InformedRO}
J.~D. Gammell, S.~S. Srinivasa, and T.~D. Barfoot, ``Informed {RRT}*: Optimal
  sampling-based path planning focused via direct sampling of an admissible
  ellipsoidal heuristic,'' \emph{2014 IEEE/RSJ International Conference on
  Intelligent Robots and Systems}, pp. 2997--3004, 2014.

\bibitem{Selin_2019IEEERAL}
M.~Selin, M.~Tiger, D.~Duberg, F.~Heintz, and P.~Jensfelt, ``Efficient
  autonomous exploration planning of large-scale 3-{D} environments,''
  \emph{IEEE Robotics and Automation Letters}, vol.~4, no.~2, pp. 1699--1706,
  2019.

\bibitem{Oleynikova_2018IEEERAL}
H.~Oleynikova, Z.~Taylor, R.~Siegwart, and J.~Nieto, ``Safe local exploration
  for replanning in cluttered unknown environments for microaerial vehicles,''
  \emph{IEEE Robotics and Automation Letters}, vol.~3, no.~3, pp. 1474--1481,
  2018.

\bibitem{Furrer2016Rotors}
F.~Furrer, M.~Burri, M.~Achtelik, and R.~Siegwart, \emph{Robot Operating System
  (ROS): The Complete Reference (Volume 1)}.\hskip 1em plus 0.5em minus
  0.4em\relax Cham: Springer International Publishing, 2016, ch. RotorS---A
  Modular Gazebo MAV Simulator Framework, pp. 595--625.

\bibitem{Oleynikova_2017IROS}
H.~Oleynikova, Z.~Taylor, M.~Fehr, R.~Siegwart, and J.~Nieto, ``Voxblox:
  Incremental {3D} {E}uclidean signed distance fields for on-board {MAV}
  planning,'' in \emph{IEEE/RSJ International Conference on Intelligent Robots
  and Systems (IROS)}, 2017,
  \url{https://github.com/ethz-asl/mav_voxblox_planning}, accessed Aug. 2019.

\bibitem{Cieslewski_2017IROS}
T.~Cieslewski, E.~Kaufmann, and D.~Scaramuzza, ``Rapid exploration with
  multi-rotors: A frontier selection method for high speed flight,'' in
  \emph{2017 IEEE/RSJ International Conference on Intelligent Robots and
  Systems (IROS)}.\hskip 1em plus 0.5em minus 0.4em\relax IEEE, 2017, pp.
  2135--2142.

\end{thebibliography}
\end{document}